\documentclass[runningheads]{llncs}
\usepackage{graphicx}
\usepackage{amsmath,amssymb,amsfonts}
\usepackage{wrapfig}
\usepackage{xcolor}
\usepackage{enumitem}

\newtheorem{defi}{Definition}

\newtheorem{assumption}{Assumption}
\newcommand{\sig}[1]{{\small\textsf{{#1}}}}
\newcommand{\Comment}[1]{}

\begin{document}
\title{Safety-Aware Hardening of 3D Object Detection Neural Network Systems}
\author{Chih-Hong Cheng}
\authorrunning{Chih-Hong Cheng}
\institute{DENSO AUTOMOTIVE Deutschland GmbH\\
Freisinger Str. 21, 85386 Eching, Germany\\
\email{c.cheng@denso-auto.de}}
\maketitle              % typeset the header of the contribution

\vspace{-5mm}
\begin{abstract} We study how state-of-the-art neural networks for 3D object detection using a single-stage pipeline can be made safety aware. We start with the safety specification (reflecting the capability of other components) that partitions the 3D input space by criticality, where the \emph{critical area} employs a separate criterion on robustness under perturbation, quality of bounding boxes, and the tolerance over false negatives demonstrated on the training set. In the architecture design, we consider symbolic error propagation to allow feature-level perturbation. Subsequently, we introduce a specialized loss function reflecting (1) the safety specification, (2) the use of single-stage detection architecture, and finally, (3) the characterization of robustness under perturbation. We also replace the commonly seen non-max-suppression post-processing algorithm by a safety-aware \emph{non-max-inclusion} algorithm, in order to maintain the safety claim created by the neural network. The concept is detailed by extending the state-of-the-art \textsf{PIXOR} detector which creates object bounding boxes in bird's eye view with inputs from point clouds. 

%\keywords{neural networks \and \sig{PIXOR} \and provable robustness \and safety argumentation \and theory}
\end{abstract}

\vspace{-10mm}
\section{Introduction}

For perceiving the environment in automated driving, techniques for detecting object presence in 3D have been predominantly implemented using deep neural networks. While state-the-the-art implementations for 3D object detection achieve superior performance, for building trust over the created artifact, the underlying engineering process should be \emph{safety-aware}. Concretely, for certification authorities, it is essential to demonstrate that the safety specification is reflected in the design of the neural network and is aligned with the design of the post-processing algorithm.

In this paper, we study how safety concepts can be integrated into engineering 3D object detection networks with single-stage detection. Our process starts by defining the \emph{critical area} and the associated quality attributes. Intuitively, the critical area is the area nearby the ego vehicle where failed detection of an object may lead to immediate safety risks. How one defines the critical area can be dependent on the capability of the ego vehicle, such as maximum braking forces. Apart from detecting the presence of objects, the quality of detection should also be dependent on the safety characteristics of other components in the system such as motion planners. For instance, some collision avoidance algorithms may assign a fixed buffer around the bounding boxes (provided by the perception module as input) to perform planning. If the size of the real object exceeds the predicted bounding box by the buffer size, physical collisions, although unaware by motion planners, may appear. The separation of the critical area and the non-critical area leads to the separation of the corresponding architectural design and the definition of the loss function, where the subsequent design is \emph{guided by safety} for the critical area and but is \emph{driven by performance} for the non-critical area. We then restrict ourselves to the training and the post-processing pipeline, and exemplify how a single-stage detection network such as \textsf{PIXOR}~\cite{yang2018pixor} and its post-processing algorithm should be altered. For prediction in the critical area, we adapt training techniques that incorporate feature-level perturbation, and define a loss function over perturbed worst case, the label, and the allowed tolerance. We also provide a mathematically proven bound to associate (i) the tolerance in prediction-label difference and (ii) the required buffer to ensure that a prediction contains the labeled bounding box. For post-processing, we use an alternative \emph{non-max-inclusion} algorithm, which intuitively enlarges the predicted bounding box by considering other predictions on the same object with a slightly lower probability. The use of non-max-inclusion is conservative, but it ensures that the safety claims made in the neural network remain valid after post-processing. 

By including this concept to the development process, one immediate benefit is for certification authorities to obtain profound transparency regarding how safety concepts can be demonstrated in the concrete design of neural networks and the following post-processing algorithm. Our proposal does not prohibit the use of standard training with commonly seen loss functions (e.g., mean square error). Instead, one may use parameters trained with standard approaches  as initializing parameters; subsequently, perform parameter fine-tuning under special loss functions that consider area-criticality and robustness. Our proposal also does not prohibit the use of other further post-processing algorithms such as utilizing time-series data, as in single-stage detection networks, non-max-suppression is executed immediately on the output of the neural network, and our proposal merely replaces non-max-suppression by a customized algorithm. 

In summary, our primary contributions are (i) an exemplification of the safety concept reflected into the architecture design and the corresponding post-processing; (ii) a formally stated constraint associating the quality of the prediction and the its effect on the interacting components such as planners; and (iii) the extension of provable robustness into single-stage object detection networks.\footnote{Due to space limits, we refer readers to the appendix for proofs of the lemmas and our preliminary evaluation. }

%This rest of the paper is organized as follows. After summarizing related work in Section~\ref{sec.related}, Section~\ref{sec.nn.and.pixor} introduces the terminology and the \sig{PIXOR} architecture. Section~\ref{sec.safety.goal.arguments} enumerates the safety goal and the proposed safety arguments. Reflecting on the safety arguments,  Section~\ref{sec.architecture.design} presents the architecture design, the new loss function, and the conservative post-processing algorithm. Finally,  Section~\ref{sec.concluding.remarks} concludes with further directions. Due to space limits, we refer readers to the appendix for the evaluation and proofs of the lemmas. 

\vspace{-2mm}
\section{Related Work}~\label{sec.related}
\vspace{-4mm}

There exist many novel neural network architectures that can perform object detection from point clouds, inlcuding \sig{PIXOR}~\cite{yang2018pixor}, \sig{VeloFCN}~\cite{li2016vehicle}, \sig{PointRCNN}~\cite{shi2019pointrcnn}, \sig{Vote3Deep}~\cite{engelcke2017vote3deep}, among others. Our work starts with the idea of refining the \sig{PIXOR} system to make (i) the design of neurons, (ii) the design of the loss function, and (iii) the immediate post-processing algorithm linked to requirements imposed from the safety argumentation and capabilities from other components. This makes the overall system design driven by safety rather than by performance, in contrast to existing design methodologies. For example, standard \sig{PIXOR} only sets the loss to zero when the prediction perfectly matches the label. Our extension sets the loss to be zero, so long as the prediction is close to the label by a fixed tolerance. This allows the loss to be overloaded on cases with perturbation, where the perturbed input also leads to zero loss, so long if the produced output falls into the tolerated bound. Another example is to use the non-max-inclusion post-processing algorithm that enables to maintain provable guarantees in contrast to the standard non-max-suppression algorithm. 

For engineering robust neural networks, the concept in this paper is highly related to the work of provably robust training~\cite{kolter2017provable,sinha2017certifiable,wang2018mixtrain,raghunathan2018certified,wong2018scaling,tsuzuku2018lipschitz,salman2019provably}. While current research results target simple classification or regression tasks, our focus is to extend these results such that the technique scalably applies to single-stage object detection neural networks that produce a vector of outputs on each grid. 

Lastly, we are also aware of fruitful research results in safety certification of machine learning components, with some focusing on safety argumentation~\cite{burton2019confidence,gauerhof2018structuring,koopman2019credible,matsuno2019tackling} while others on testing and formal verification (see~\cite{huang2018survey} for a survey on results in formal verification, testing, and adversarial perturbation). While these works contribute to the overall vision of rigorous safety engineering of neural networks, these results do not touch the construction of the neural networks. In this way, our research well complements these results as we focus on constructing the neural network (architecture and the loss function) and the immediate post-processing algorithm such that they can  reflect the safety argument.

\vspace{-2mm}
\section{Neural Networks and the \textsf{PIXOR} Architecture}~\label{sec.nn.and.pixor}
\vspace{-4mm}

A deep neural network is comprised of $N$ layers
where operationally,  the $n$-th layer for $n\in\{1,\dots,N\}$ of the network is a function $g^{(n)}: \mathbb{R}^{d_{n-1}} \rightarrow \mathbb{R}^{d_{n}}$, with $d_{n}$ being the dimension of layer~$n$.  
Given an input $\sig{in} \in \mathbb{R}^{d_{0}}$, the output of the $n$-th layer of the neural network $f^{(n)}$ is given by the functional composition of the $n$-th layer and the previous layers $f^{(n)}(\sig{in}) := \circ_{i=1}^{(n)} g^{(i)}(\sig{in})  = g^{(n)}(g^{(n-1)}\ldots g^{(2)}(g^{(1)}(\sig{in})))$. Given input~$\sig{in}$, the prediction of the neural network is thus $f^{(N)}(\sig{in})$. We use $g^{(n)}_j$ to denote the $j$-th neuron in layer~$n$, and for fully connected layers, computing $g^{(n)}_j$ is done by a weighted sum (characterized by the weight matrix $W^{n}_{j}$ and bias $b_{j}$) followed by applying a nonlinear activation function $\rho^{n}_j$, i.e., $g^{(n)}_j := \rho^{n}_j ((\sum^{d_{(n-1)}}_{k=1} W^{n}_{jk}g^{(n-1)}_{k}(\sig{in})) + b_{j})$. Note that such a definition allows to represent commonly seen architecture variations such as residual blocks or top-down branches that are used in \textsf{PIXOR}.
%\footnote{Architectural approaches such as residual blocks or top-down branches may feed a particular neural network layer~$k$ with inputs from the output of layer~$k'$ where $k' < k-1$. To represent this to our formulation, the output from the~$k'$-th layer can be wrapped with a sequence of identity functions until layer~$k$.}. 
Given a tensor~$v$, we use subscript~$v_{\langle i,j\rangle}$ to extract the tensor by taking the $i$-th element in the first dimension and $j$-th element in the second dimension. Lastly, define the \emph{training set} for the neural network to be $\mathcal{T} := \{(\sig{in} , \sig{lb})\}$ where for every input $\sig{in} \in \mathbb{R}^{d_{0}}$, $\sig{lb} \in \mathbb{R}^{d_{N}}$ is the corresponding label.

\textsf{PIXOR}~\cite{yang2018pixor}  is a \emph{single-stage detector} neural network that, based on the input as point clouds, directly predicts the final position and orientation of an object. By single-stage, we refer to neural networks where there exists no intermediate step of proposing possible areas (region proposal) in 3D for suggesting the potential existence of an object in the area. In this paper, we follow the original formulation in~\cite{yang2018pixor} to only detect the presence of a ``car" as in the KITTI dataset~\cite{geiger2013vision}. One can easily extend the network architecture to include other object types.  
%\begin{figure}[t]
\begin{wrapfigure}{r}{0.5\textwidth}
\centering
\vspace{-30pt}
\includegraphics[width=0.5\textwidth]{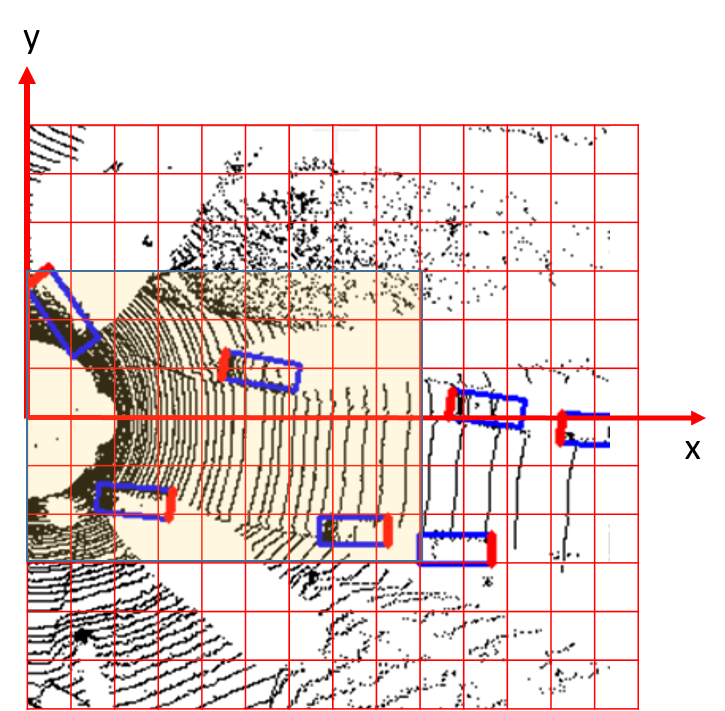}
\caption{Visualizing the relation between grids  and $x$-$y$ coordinates. The bottom-left grid has an index $\langle 0,0 \rangle $, and its center point is translated to the $x$-$y$ coordinate equaling $(\frac{\alpha}{2}, -\frac{11\alpha}{2})$.}
\vspace{-10pt}
\label{fig:grid}
\end{wrapfigure}
%\end{figure}

The input dimension $d_{0}$ of the \textsf{PIXOR} network is $(L,W,H)$. Provided that the positive direction of the~$x$ coordinate  facing in the front windshield of the car, the positive direction of the~$y$ coordinate facing to the left of driver, and the positive direction of the~$z$ coordinate
facing up, each $\sig{in}_{\langle i,j,k\rangle}$ contains the density of the lidar point cloud centered at the point $(i\alpha + \frac{\alpha}{2}, (j-\frac{W}{2})\alpha + \frac{\alpha}{2}, k\alpha + \frac{\alpha}{2})$ with $\alpha$ being the size of the grid. Figure~\ref{fig:grid} illustrates how grids are mapped to the physical 2D dimension.

\begin{figure}[t]
\centering
\includegraphics[width=0.7\textwidth]{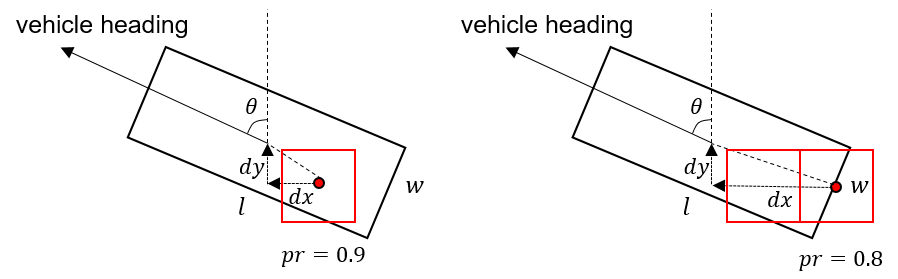}
\vspace{-5mm}
\caption{The output of the \textsf{PIXOR} network for a single grid; the red dot represents the centering position of the grid. Both two grids have positive prediction on the existence of the same vehicle. }
\label{fig:output}
\end{figure}

The output dimension for the network is $d_{L} = (\frac{L}{\beta},\frac{W}{\beta}, 7)$ with $\beta$ being the down-scaling factor. The output $f^{(N)}(\sig{in})_{\langle i,j \rangle}$, i.e., the output at grid $\langle i,j\rangle$, is a $7$ tuple $(pr, \cos(\theta), \sin(\theta), dx,  dy, \log(w), \log(l))$, where $\langle i,j\rangle$ is matched to the physical area in 2D centered by point $(i\alpha\beta + \frac{\alpha\beta}{2}, (j-\frac{W}{2\beta})\alpha\beta + \frac{\alpha\beta}{2})$ with $\alpha\beta$ being the size of the output grid. The output of the network essentially predicts, for each grid in $\langle i,j\rangle$ in the 2D plane, if there is a vehicle nearby. In the original \textsf{PIXOR} paper~\cite{yang2018pixor}, the grid size $\alpha$ is set to~$0.1$ meter and $\beta$ is set to~$4$, meaning that the output grid has a physical quantity of $0.4\times 0.4\;m^2$. Figure~\ref{fig:output} shows the visualization of the output relative to the center of the grid. One direct consequence is that it is possible for multiple grids to create prediction over the same vehicle (as the grid size is very small),  by having a different displacement value $dx$ and $dy$ to the center of the vehicle. Therefore, a post-processing algorithm called \emph{non-max-suppression} is introduced. The idea is to first pick the output grid whose prediction probability is the largest while being larger than the \emph{class threshold}~$\alpha$. Subsequently, create the prediction as a bounding box, and remove every prediction whose bounding box  overlaps with the previously created bounding box (for the same type) by a certain threshold, where the degree of overlapping is computed using the intersection-over-union (IoU) ratio. For the example in Figure~\ref{fig:output}, the vehicle prediction on the right will be neglected as the probability ($pr=0.8$) is less than the prediction probability from the left grid ($pr=0.9$). Note that  \emph{such a post-processing algorithm, its computation is not integrated the training and the inference of neural networks}. 

\vspace{-4mm}
\section{Safety Goals and the Proposed Safety Arguments}~\label{sec.safety.goal.arguments}
\vspace{-5mm}

    %\begin{figure}[t]
\begin{wrapfigure}{r}{0.42\textwidth}
\centering
\vspace{-20pt}
\includegraphics[width=0.4\textwidth]{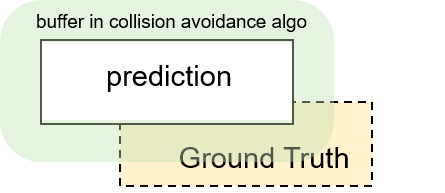}
\caption{Understanding the quality of prediction. The area of ground truth that is outside the buffer of the collision avoidance algorithm imposes risk of collision. }
    \label{fig:prediction.quality}
\end{wrapfigure}

We define the safety goal by first considering the \emph{critical area}. Intuitively, a critical area is an area where object detection should be processed with care, as detection miss or an object prediction of the wrong size may lead to unsafe consequences. The precise definition of the critical area and the associated quality attribute can be driven by multiple factors such as the specification of labeling quality, the capability of motion planners, and the capability of maximum breaking. In the following, we create the following sample specification and describe the underlying rationale.

\vspace{-2mm}
\begin{description}
    \item[(S1: Critical area)] The critical area are output grids $\langle i, j\rangle$ where $i \in [0, \gamma_L]$ and $j \in [\frac{W}{2\beta} - \gamma_W, \frac{W}{2\beta} + \gamma_W]$. Figure~\ref{fig:grid} shows an example where critical area (in light yellow) is in grid $\langle i, j\rangle$ with $i \in [0, 9]$ and $j \in [3, 8]$.

    \item[(S2: Demonstrate no false negative in critical area)]  \emph{If there exists an object in the critical area, it should be ``detected"}. The system may fail to detect  an object outside the critical area but there is no immediate danger (such as hitting an object). The meaning of ``detected" is detailed as follows: \emph{the predicted bounding box (of the vehicle) should deviate from the ground truth with a fixed tolerance}. In later sections, we provide the mathematical formulation to quantify the meaning of tolerance in prediction. 
    \begin{itemize}
        \item   The rationale is that we assume that the collision avoidance algorithm takes a fixed buffer in its planning, and it fully trusts the output of the object detection. Therefore, any prediction that deviates from the ground truth with an amount more than the tolerance can create the risk for collision (see Figure~\ref{fig:prediction.quality} for an illustration).
        \item The ``if" condition in the specification implies that safety constraint is only on false negatives (undetected objects) and is not on false positives (ghost objects). In other words, reducing false positives is a performance specification and (within the scope of this paper) is not considered as a safety specification.\footnote{This paper targets automated driving systems (SAE J3016 level~3 and up); for ADAS systems (SAE J3016 level~2) such as automatic emergency braking, false negatives are less critical due to the driver taking ultimate control, but avoiding false positives are considered to be safety-critical due to potential rear collision. } 
        %\item 
    \end{itemize}

    \item[(S3: High performance outside the critical area)] For objects outside the critical area, the prediction should achieve reasonable performance, but \emph{no hard constraints are imposed}. 
    
\end{description}

\vspace{-2mm}

Therefore, we employ two philosophies in designing the quality attributes for object detection. Within the critical area, the quality attribute is \emph{safety-driven} - performance shall never sacrifice safety. This implies that a neural network that creates tighter bounding boxes but may fail to detect an object in the critical area is not allowed. Based on the specification, the training of neural networks may reduce false positives that appeared inside the critical area, but it is only a performance improvement. Outside the critical area, the quality attribute is \emph{performance-driven} - positives and false negatives are allowed for object detection outside the critical area. However, the training of neural networks may try to reduce them. Lastly, in this example, we do not enforce perfection between the prediction given input and the associated ground truth label. As demonstrated in later sections, this is reflected by having a zero loss so long as the prediction deviates from the ground truth by the tolerance.

Following the safety specification listed above, we are considering the following safety arguments in the architecture design and post-processing to support the key safety specification~\textbf{(S2)}. Note that the listed items are partial and can be further expanded to create a stronger supporting argument.

\vspace{-2mm}
\begin{description}

    \item[(A1: Safety-aware loss function)] The first argument is a careful definition of a loss function that reflects the requirement listed above. In particular, for critical and non-critical areas, different loss functions are applied.
    \item[(A2: Robust training for critical area)] For generating predictions inside the critical area, the second argument is to deploy specialized training mechanisms such that one has a theoretical guarantee on robust prediction over data used in training, provided that the robust loss has dropped to zero.\footnote{It is almost impossible to use standard loss functions while demonstrating zero loss, as zero loss implies perfection between prediction and labels. Our robust training and our defined robust loss can, as demonstrated in later sections,  enable zero loss (subject to parameters used) in practical applications.} 
    \item[(A3: Conservative post-processing)] For post-processing algorithms that are independent of the neural network but crucial to the generated prediction, they should act more conservatively in the critical area.  

\end{description}

\begin{figure}[t]
%trim={<left> <lower> <right> <upper>}
    \centering
    \includegraphics[width=0.95\textwidth, trim=0cm 6.8cm 2cm 2cm, clip]{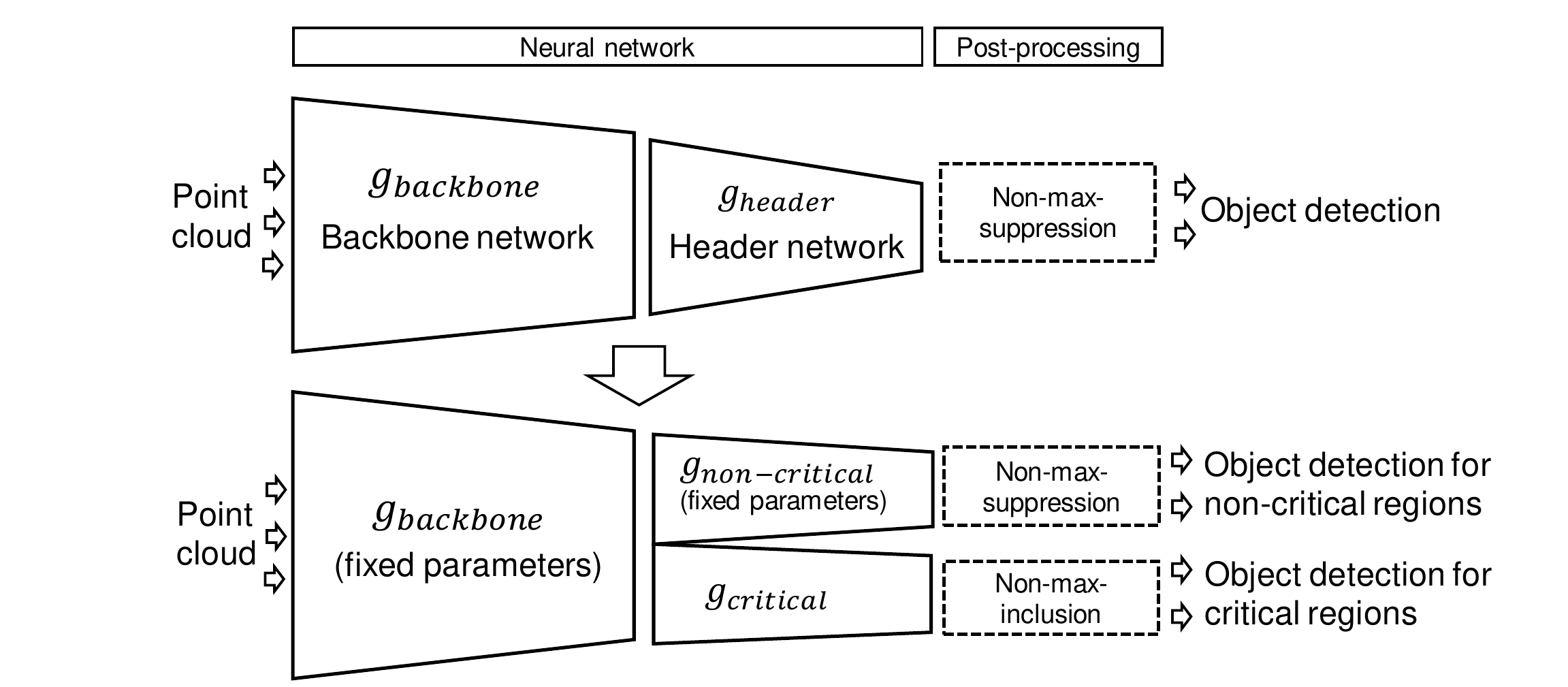}
    \caption{Hardening a standard object detection network system (top) to a safety-aware one that differentiates between critical and non-critical areas (down).}
    \label{fig:hardening}
\end{figure}

\vspace{-4mm}
\section{Architecture Design, Loss, and Post-Processing}~\label{sec.architecture.design}
\vspace{-4mm}

In this section, we detail how we extend the architecture of \textsf{PIXOR} to incorporate a safety-aware loss function, to integrate a new post-processing algorithm, and finally, to apply robust training techniques.  Recall that \textsf{PIXOR} has a \emph{backbone network} $g_{backbone}$ and a \emph{header network} $g_{header}$. Intuitively, the backbone network creates high-level features from high dimensional inputs, and the header network produces predictions from high-level features. 

We illustrate the proposed network architecture and the associated post-processing pipeline in Figure~\ref{fig:hardening}. 
We start by training a standard \textsf{PIXOR} network as a baseline model. It is tailored for optimal performance but is not safety-aware. In the hardened neural network, we create two  header networks $g_{critical}$ and $g_{non.critical}$, with both connected to the backbone network $g_{backbone}$. Parameters in $g_{backbone}$ are fixed, i.e., they are not subject to further change. $g_{non.critical}$ is used to perform object detection for the non-critical area, and $g_{critical}$ is used to perform object detection for the critical area. In this paper, we present two variations to engineer $g_{critical}$ together with their associated theoretical guarantees. 
Lastly, for the critical area, a non-standard post-processing algorithm (\emph{non-max-inclusion}) is used rather than the standard non-max-suppression algorithm which is used for non-critical areas. 

\vspace{-2mm}
\paragraph{(Engineering $g_{non.critical}$)}  For $g_{non.critical}$ to perform object detection in non-critical areas, it is created by taking $g_{header}$ and subsequently, remove all neurons in the final layer that generate predictions for critical areas. Naturally, such a header network can be used for making predictions that are outside the critical area. We do not perform further training to change any learned parameters. 
%trim={<left> <lower> <right> <upper>}
\begin{figure}[t]
    \centering
    \includegraphics[width=\textwidth, trim=3.5cm 6cm 3cm 8.5cm, clip]{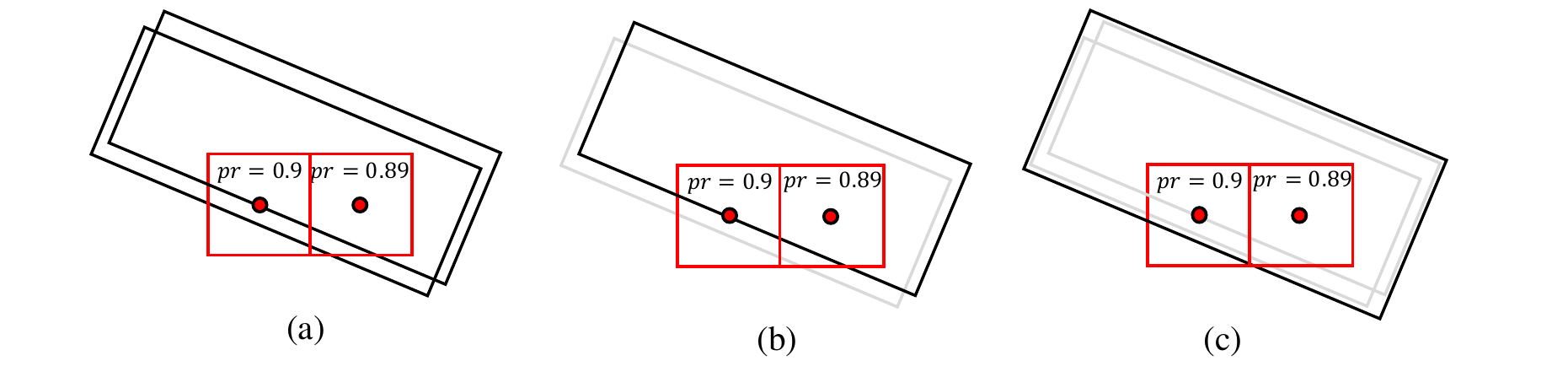}
    \vspace{-5mm}
    \caption{Explaining non-max-inclusion.}
    \label{fig:non.max.inclusion}
\end{figure}

\vspace{-2mm}
\subsection{Post-processing algorithm for the critical area (addressing \textbf{A3})} 

For the post-processing in the critical area,  we alternatively propose the \emph{non-max-inclusion} algorithm. The underlying idea is illustrated using Figure~\ref{fig:non.max.inclusion}, where a \sig{PIXOR} network generates the prediction of the same object on two adjacent grids. Standard non-max-suppression (Figure~\ref{fig:non.max.inclusion}-b) picks the grid that produces the largest prediction probability and suppresses other predictions with high intersection-over-union (IoU); in Figure~\ref{fig:non.max.inclusion}-b the bounding box with probability equaling~$0.89$ is suppressed. Nevertheless, as there is no ground truth in operation, it is uncertain which bounding box is correct. Therefore, a conservative post-processing algorithm should include both bounding boxes, as demonstrated in Figure~\ref{fig:non.max.inclusion}-c. Overall, the non-max-inclusion algorithm proceeds as follows.
\begin{enumerate}
    \item Sort all bounding boxes with their prediction probabilities (from high to low), and store them to a list $L_{all}$. Remove all boxes whose probability is lower than the class threshold~$\alpha$.
    \item Remove from $L_{all}$ the first bounding box $B$ (i.e., the box with the highest probability), and create a sub-list $L_{B} =\{B\}$ containing only $B$. 

    \item For each bounding box prediction $B' \in L_{all}$, if the prediction $B'$ has high IoU with the original bounding box $B$, remove $B'$ from $L_{all}$ and add~$B'$ to~$L_{B}$ and do not consider it afterwards.
    
    \item Finally, build the final bounding box~$\hat{B}$ which contains all bounding boxes of~$L_{B}$ and use it as the final prediction for $B$.
    
    \item   Proceed to step~2 until $L_{all}$ is empty. 
\end{enumerate}

\vspace{-5mm}
\subsection{Engineering $g_{critical}$ - variation 1 (addressing \textbf{A1})}
%\vspace{-2mm}

In the following, we introduce the first variation for engineering  $g_{critical}$ addressing (\textbf{A2}), where we  create $g_{critical}$ by first taking $g_{header}$, followed by removing all connections to final layer neurons that generate predictions for non-critical areas.  In contrast to $g_{non.critical}$, the parameters for weights and bias (from $g_{header}$) will be further adjusted due to the newly introduced loss function.   

\paragraph{(Loss function inside the critical area)} For a labeled data $(\sig{in}, \sig{lb}) \in \mathcal{T}$, we need to define the loss for every output grid $\langle i, j\rangle$ that is inside the critical area, and the overall loss is the sum of loss from each grid. In particular, the loss function shall reflect the safety specification:

\begin{itemize}
    \item If in the ground truth, there exists a vehicle centered at output grid $\langle i, j\rangle$, then the loss for output grid $\langle i, j\rangle$ is~$0$ so long as 
    \begin{itemize}
        \item for the 1st output prediction, i.e., the predicted probability, it is greater than \emph{class threshold}~$\alpha$, as the post-processing algorithm uses~$\alpha$ as the threshold, and     
       \item  for the $k$-th output prediction where $k > 1$, the distance between the output and the label is bounded by $\delta_k$.
    \end{itemize}
    
    \item If in the ground truth, there does not exist a vehicle centered  at output grid $\langle i, j\rangle$, then loss for output grid $\langle i, j\rangle$ is~$0$ so long as 
    \begin{itemize}
        \item the prediction probability is less than \emph{class threshold}~$\alpha$, as for the post-processing algorithm, it uses~$\alpha$ as the threshold.
    \end{itemize} 
    
\end{itemize}

\begin{defi}\label{defi.distance}
Let function $\sig{dist}(v, \sig{low}, \sig{up})$ return the minimum distance between value $v$ to the interval $[\sig{low}, \sig{up}]$. That is, $\sig{dist}(v, \sig{low}, \sig{up}) := \sig{min}(|v- \sig{low}|, |v-\sig{up}|)$. 
\end{defi}

\noindent The following definition of loss captures the above mentioned concept. 

\begin{defi}\label{defi.loss}
Given $(\sig{in},\sig{lb}) \in \mathcal{T}$, define  the loss between the prediction $\sig{out} := f^{(n)}(\sig{in})$ and the ground truth $\sig{lb}$ at output grid ${\langle i, j\rangle}$ to be  

\vspace{-3mm}

\begin{equation*}
\sig{loss}_{\langle i, j\rangle}(\sig{out}, \sig{lb}) := 
\begin{cases}
%1 &\text{se $\omega\in A$}\\
%1250 &\text{se $\omega \in A^c$}
\sig{dist}(\sig{out}_{\langle i, j, 1\rangle}, \alpha, \infty) + \eta & \text{if $\sig{lb}_{\langle i, j, 1\rangle} = 1$} \\
\sig{dist}(\sig{out}_{\langle i, j, 1\rangle}, -\infty,  \alpha) + \eta   & \text{otherwise $(\sig{lb}_{\langle i, j, 1\rangle} = 0)$} 
\end{cases}
\end{equation*}

where $\eta := \sum^{7}_{k=2} \sig{dist}(\sig{out}_{\langle i, j, k\rangle}, \sig{lb}_{\langle i, j, k\rangle} - \delta_k,  \sig{lb}_{\langle i, j, k\rangle} + \delta_k)$
\end{defi}

\noindent At the end of this section, we provide technical details on how to implement the loss function using state-of-the-art machine learning framework PyTorch.

\vspace{-2mm}
\paragraph{(Connecting zero loss with buffer size)} To avoid scenarios shown in Figure~\ref{fig:prediction.quality}, the following lemma provides a conservative method to enlarge the predicted bounding box, such that the enlarged bounding box guarantees to contain the ground truth bounding box so long as the computed loss value $\sig{loss}_{\langle i, j\rangle}(\sig{out}, \sig{lb})$ equals~$0$.  

\begin{lemma}\label{lemma.buffer.bound} 

For the labelled data $(\sig{in}, \sig{lb}) \in \mathcal{T}$, given $\sig{out} := f^{(n)}(\sig{in})$, let $\theta$ be the angle produced by $\sig{out}_{\langle i, j, 2\rangle}$ and $\sig{out}_{\langle i, j, 3\rangle}$. If $\sig{loss}_{\langle i, j\rangle}(\sig{out}, \sig{lb}) = 0$ and if $\sig{lb}_{_{\langle i, j, 1\rangle}} = 1$, then enlarging the prediction bounding box by 

\vspace{-2mm}
\begin{itemize}
    \item enlarging the predicted width with $d + d_w$
    \item enlarging the predicted length with $d + d_l$
\end{itemize}
\vspace{-2mm}
guarantees to contain the vehicle bounding box from label $\sig{lb}$ at grid ${\langle i, j\rangle}$, where $d$, $d_l$ and $d_w$ need to satisfy the following:

\vspace{-2mm}
\begin{itemize}
    \item $d > \sqrt{(\delta_4)^2 + (\delta_5)^2}$,

    \item  $d_l > \sig{max}_{\alpha \in [-\kappa, \kappa]}\;\frac{1}{2}(10^{\sig{out}_{\langle i, j, 6\rangle} + \delta_6} \sin \alpha + 10^{\sig{out}_{\langle i, j, 7\rangle} + \delta_7} \cos \alpha - 10^{\sig{out}_{\langle i, j, 7\rangle}}) $, 

\item  $d_w > \sig{max}_{\alpha \in [-\kappa, \kappa]}\;\frac{1}{2}(10^{\sig{out}_{\langle i, j, 6\rangle} + \delta_6} \cos \alpha + 10^{\sig{out}_{\langle i, j, 7\rangle} + \delta_7} \sin \alpha - 10^{\sig{out}_{\langle i, j, 6\rangle}})$,
\end{itemize}

\vspace{-2mm}
and the interval $[-\kappa, \kappa]$ used in $d_l$ and $d_w$ satisfies the following constraints: 

\vspace{-8mm}
\begin{multline}\label{eq.enlarge.buffer}
\forall \alpha \in [\frac{-\pi}{2},\frac{\pi}{2}]: (|\cos (\theta ) - \cos (\theta+\alpha)| \leq \delta_2 \; \wedge   |\sin (\theta ) - \sin (\theta+\alpha)| \leq \delta_3) \\ \rightarrow  \alpha \in [-\kappa, \kappa] 
\end{multline}
\end{lemma}

A conservative computation of $d_l$ and $d_w$ independent of the generated prediction can be done by further assuming the maximum length and width of a vehicle for the observed output, e.g., a vehicle can have at most $6$ meters in length ($10^{\sig{out}_{\langle i, j, 7\rangle}} \leq 6$) and $2.5$ meters in width ($10^{\sig{out}_{\langle i, j, 6\rangle}} \leq 2.5$); the assumptions shall be monitored in run-time to check if one encounters prediction that generates larger vehicle bounding boxes. As $\delta_2, \ldots, \delta_7$ are constants, and by setting $[-\kappa, \kappa]$ to be a constant interval (which is related to the value of $\delta_2$ and $\delta_3$)\footnote{The interval $[-\kappa, \kappa]$ is essentially a \emph{conservative upper bound on the deviated angle} between prediction and the ground truth, where their associated sine and cosine value differences are bounded by $\delta_2$ and $\delta_3$. As an example, if between  the predicted angle and the ground truth, we only allow the sine and cosine value to only differ by at most $0.1$ (i.e., $\delta_2=\delta_3=0.1$), it is easy to derive that $\kappa$ can be conservatively set to~$\frac{\pi}{18}$, i.e., ($10^{\circ}$), rather than the trivial value~$\frac{\pi}{2}$. This is because an angle difference of $10^{\circ}$ can already make sine and cosine value differ by~$0.15$, thereby creating non-zero loss. Therefore, conservatively setting $[-\kappa, \kappa]$ to be $[-\frac{\pi}{18}, \frac{\pi}{18}]$ in computing $d_l$ and $d_w$ surely covers all possible angle deviation constrained by zero loss.}, the minimum value for $d_l$ and $d_w$ can be computed using numerical approximation solvers such as Mathematica or Sage.  

%\vspace{2mm}

\vspace{-2mm}
\paragraph{(Connecting the prediction and the post-processing algorithm)} As a consequence, given $(\sig{in}, \sig{lb}) \in \mathcal{T}$, provided that $\sig{loss}_{\langle i, j\rangle}(\sig{out}, \sig{lb}) = 0$ and $\sig{lb}_{{\langle i, j, 1\rangle}} = 1$, one ensures the following:
\begin{enumerate}
    \item By enlarging the bounding box created from the prediction using Lemma~\ref{lemma.buffer.bound},  the enlarged bounding box is guaranteed to contain the bounding box created by the label.
    \item The non-max-inclusion algorithm never removes any bounding box with prediction probability smaller than $\alpha$, 
\end{enumerate}
Therefore, the resulting list of bounding boxes after post-processing is \emph{guaranteed to have one bounding box that completely contains the ground truth}. This implies that situation in Figure~\ref{fig:prediction.quality} does not occur in~$(\sig{in}, \sig{lb})$.

\paragraph{(Problems in using standard post-processing algorithm)} Notice that the above mentioned guarantee \emph{does not hold when replacing non-max-inclusion with non-max-suppression}, as the enlarged bounding box from the prediction at $\langle i, j\rangle$ (which has the guarantee of containing the ground truth) can be removed, so long as there exists another bounding box from a nearby output grid $\langle i', j'\rangle$ that (i) has higher predicted probability value (i.e., $\sig{out}_{\langle i', j', 1\rangle} \geq \sig{out}_{\langle i, j, 1\rangle}$) and (ii) has a huge area overlap with the one from grid $\langle i, j\rangle$. 

\vspace{-2mm}
\subsection{Engineering $g_{critical}$ - variation 2  (addressing \textbf{A1} and \textbf{A2})}

\begin{figure}[t]
%trim={<left> <lower> <right> <upper>}
    \centering
    \includegraphics[width=0.9\textwidth, trim=0.5cm 1.5cm 0.5cm 0.5cm, clip]{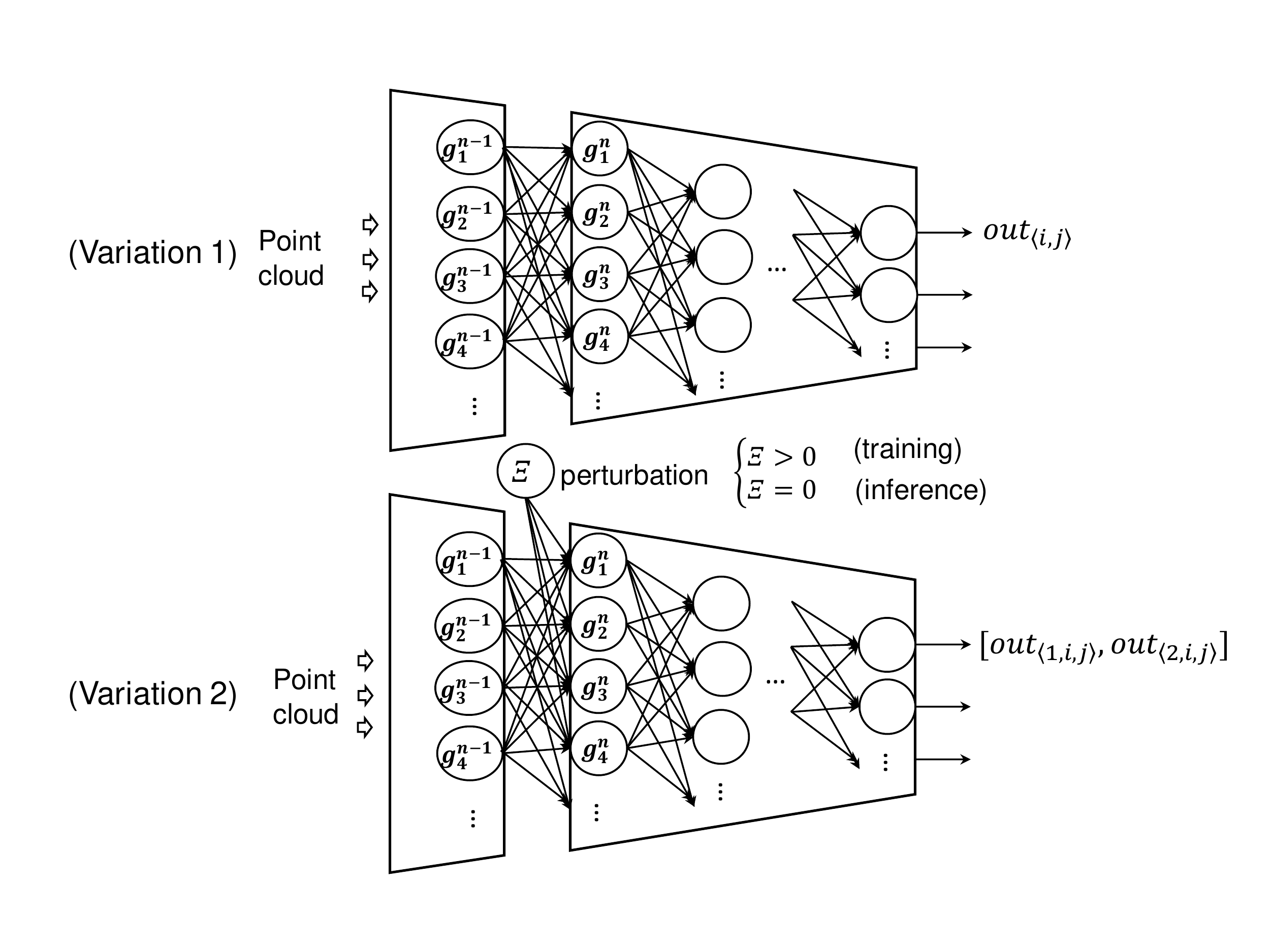}
    \caption{Introducing feature-level perturbation in engineering $g_{critical}$.}
    \label{fig:symbolic.feature.loss}
\end{figure}

We propose an improvement for engineering $g_{critical}$ which which considers feature-level robustness, thereby also addressing~\textbf{A2}. The underlying idea is illustrated in Figure~\ref{fig:symbolic.feature.loss}, where parameters to be learned are the same between the first variation and the second variation. The difference lies in how values are propagated in training (value propagation in variation~1 versus bound propagation in variation~2) and in how the loss is computed (loss accounting tolerance in variation~1 versus symbolic loss accounting tolerance in variation~2).

%\paragraph{(Architecture Design)} 

\begin{itemize}
    \item (First layer in $g_{critical}$) For every neuron in $g_{critical}$ receiving values produced from the backbone network, we require that it takes an additional input parameter~$\Xi \geq 0$ that characterizes the maximum amount of \emph{perturbation}, and subsequently computes a conservative bound accounting all possible perturbation. Let~$n$ be the starting layer index of $g_{critical}$. During training, we require neuron $g^{n}_{j}$ to compute, subject to the condition ${-\Xi \leq \xi_1, \ldots, \xi_{d_{(n-1)}} \leq \Xi}$, values $\sig{lowerb}^{n}_{j}$ and $\sig{upperb}^{n}_{j}$ where
    \begin{itemize}
        \item $\sig{lowerb}^{n}_{j} \leq \sig{min} \; \rho^{n}_j (\sum^{d_{(n-1)}}_{i=1 } W^{n}_{ji}(g^{(n-1)}_{i}(\sig{in}) + \xi_i) + b^{n}_{j})$ and 
        \item $\sig{upperb}^{n}_{j} \geq \sig{max} \; \rho^{n}_j (\sum^{d_{(n-1)}}_{i=1 } W^{n}_{ji}(g^{(n-1)}_{i}(\sig{in}) + \xi_i) + b^{n}_{j})$. 
    \end{itemize}

    \vspace{1mm}
    In other words, the interval $[\sig{lowerb}^{n}_{j}, \sig{upperb}^{n}_{j}]$ acts as a \emph{sound over-approximation} over all possible computed values by taking input $g^{(n-1)}_{i}(\sig{in})$ and by having it boundedly perturbed. 
    
        \vspace{1mm}
    \item (Other layers in $g_{critical}$) For other layers, during training, a neuron takes the bound computed by the previous layer and computes again a sound over-approximation over possible outputs. In other words, during training, the $j$-th neuron at layer~$n'$ $(n'> n)$ computes~$\sig{lowerb}^{n'}_{j}$ and~$\sig{upperb}^{n'}_{j}$ such that 

    \begin{itemize}
        \item $\sig{lowerb}^{n'}_{j} \leq \sig{min} \; \rho^{n'}_j (\sum^{d_{(n'-1)}}_{i=1 } W^{n}_{ji} \,\sig{v}^{(n'-1)}_{i} + b^{n'}_{j})$ and 
        \item $\sig{upperb}^{n'}_{j} \geq \sig{max} \, \rho^{n'}_j (\sum^{d_{(n'-1)}}_{i=1 } W^{n}_{ji}\; \sig{v}^{(n'-1)}_{i}  + b^{n'}_{j})$, 
        
    \end{itemize}

    where for $i \in 1, \ldots, d_{(n'-1)}$, $\sig{lowerb}^{n'-1}_{i} \leq \sig{v}^{(n'-1)}_{i} \leq \sig{upperb}^{n'-1}_{i}$. 

\end{itemize}

%During inference, the computation follows standard approaches where perturbation is set to~$0$, making the computed upper-bound and the lower-bound equal. 

Here we omit technical details, but  one can implement above mentioned bound computation in state-of-the-art ML training framework using abstract interpretation techniques such as dataflow analysis~\cite{cousot1977abstract}. Using dataflow analysis, under $\Xi = 0$ the computation turns exact, meaning that the lower-bound and the upper-bound should collide, i.e., $\sig{lowerb}^{n'}_{j} = \sig{upperb}^{n'}_{j}$.  In the rest of the paper, we always assume that Assumption~\ref{assumption.exact} holds. This implies that during inference, one may set~$\Xi$ to~$0$ and can use the computed lower-bound as the prediction.

\vspace{-1mm}
\begin{assumption}~\label{assumption.exact}
Assume that when $\Xi=0$, the computation of $\sig{lowerb}^{n}_{j}$, $\sig{upperb}^{n}_{j}$, $\sig{lowerb}^{n'}_{j}$, $\sig{upperb}^{n'}_{j}$ for every input $\sig{in} \in \mathbb{R}^{d_0}$ is exact, i.e.,  $\sig{lowerb}^{n}_{j} = \sig{upperb}^{n}_{j}$ and  $\sig{lowerb}^{n'}_{j} = \sig{upperb}^{n'}_{j}$.
\end{assumption}
\vspace{-4mm}

\paragraph{(Characterizing Robust Loss)} During training, as each output is no longer a single value but a bound incorporating the effect of perturbation, the loss function should be adjusted accordingly. For simplifying the notation, for each output at grid indexed $\langle i, j\rangle$, we add another dimension in the front to indicate the lower and the upper bound. That is, use $\sig{out}_{\langle 1, i, j\rangle}$ and $\sig{out}_{\langle 2, i, j\rangle}$ for indicating the computed lower- and upper-bound at grid $\langle i, j\rangle$.
 
\vspace{-2mm}
\begin{defi}\label{defi.robust.loss}
At output grid ${\langle i, j\rangle}$, define the robust loss between the range of possible values $\sig{out}$ (computed using perturbation bound $\Xi$ and the input $\sig{in}$) and the ground truth $\sig{lb}$ to be  

\vspace{-4mm}
\begin{equation*}
\sig{robust\_loss}_{\langle i, j\rangle}(\sig{out}, \sig{lb}) := 
\begin{cases}
%1 &\text{se $\omega\in A$}\\
%1250 &\text{se $\omega \in A^c$}
\sig{dist}(\sig{out}_{\langle 1, i, j, 1\rangle}, \alpha, \infty) + \frac{\eta_l + \eta_u}{2} & \text{if $\sig{lb}_{\langle i, j, 1\rangle} = 1$} \\
\sig{dist}(\sig{out}_{\langle 2, i, j, 1\rangle}, -\infty,  \alpha) +  \frac{\eta_l + \eta_u}{2}   & \text{otherwise $(\sig{lb}_{\langle i, j, 1\rangle} = 0)$} 
\end{cases}
\end{equation*}
\vspace{-1mm}
where 
\vspace{-1mm}
\begin{itemize}
    \item $\eta_l := \sum^{7}_{k=2} \sig{dist}(\sig{out}_{\langle 1, i, j, k\rangle}, \sig{lb}_{\langle i, j, k\rangle} - \delta_k,  \sig{lb}_{\langle i, j, k\rangle} + \delta_k)$ and
    \item  $\eta_u := \sum^{7}_{k=2} \sig{dist}(\sig{out}_{\langle 2, i, j, k\rangle}, \sig{lb}_{\langle i, j, k\rangle} - \delta_k,  \sig{lb}_{\langle i, j, k\rangle} + \delta_k)$
\end{itemize}

\end{defi}

The rationale for the design of robust loss is as follows: If the ground truth indicates that there exists an object at grid $\langle  i, j\rangle$, then we hope that any input under perturbation should still reports the existence of that object. This is characterized by the probability lower-bound $\sig{out}_{\langle 1, i, j, 1\rangle}$ being larger than the class threshold~$\alpha$. On the other hand, if the ground truth states that no object exists at grid $\langle  i, j\rangle$, then all possible input perturbation can report absence, so long when the probability upper-bound $\sig{out}_{\langle 2, i, j, 1\rangle}$ is less than the class threshold~$\alpha$. Finally, when $\eta_l$ and $\eta_u$ are both $0$, both the lower-bound and the upper-bound due to perturbation are within tolerance. In other words, the perturbation never leads to a prediction that exceeds the label by tolerance. 

Under Assumption~\ref{assumption.exact}, $\sig{robust\_loss}_{\langle i, j\rangle}(\sig{out}, \sig{lb})$ can be viewed as a generalization of $\sig{loss}_{\langle i, j\rangle}(\sig{out}, \sig{lb})$: When 
 $\Xi = 0$, computing  $\sig{robust\_loss}_{\langle i, j\rangle}(\sig{out}, \sig{lb})$
 essentially computes the same value of $\sig{loss}_{\langle i, j\rangle}(\sig{out}, \sig{lb})$ due to the colliding lower- and upper-bounds  $(\sig{out}_{\langle 1, i, j, k\rangle} = \sig{out}_{\langle 2, i, j, k\rangle})$ making $\eta_l = \eta_u$. 

Finally, given $(\sig{in}, \sig{lb}) \in \mathcal{T}$, Lemma~\ref{lemma.symbolic} (implicitly yet mathematically) characterizes the allowed perturbation on $\sig{in}$ to maintain the prediction under tolerance. If the symbolic loss at grid~${\langle i, j\rangle}$ equals zero and the label indicates the existence of an object at grid~${\langle i, j\rangle}$, then any input $\sig{in}'$ that are close to $\sig{in}$ (subject to Eq.~\ref{eq.distance.main}) should also  positively predict the existence of an object (i.e., the first output should be larger than~$\alpha$). The result can also be combined with Lemma~\ref{lemma.buffer.bound} to avoid the situation illustrated in Figure~\ref{fig:prediction.quality}.

 \vspace{-2mm}
    
\begin{lemma}~\label{lemma.symbolic}
Given $(\sig{in}, \sig{lb}) \in \mathcal{T}$, under the condition $\sig{robust\_loss}_{\langle i, j\rangle}(\sig{out}, \sig{lb}) = 0$ where $\sig{out}$ is computed using~$\sig{in}$ with perturbation bound $\Xi := \xi $ ($\xi> 0$) at layer~$n$, then for any $\sig{in}'$ where the following condition holds,

\vspace{-2mm}
\begin{equation}~\label{eq.distance.main}
   \forall m \in \{1, \ldots, d_{(n-1)}\}: |g^{(n-1)}_{m}(\sig{in'}) - g^{(n-1)}_{m}(\sig{in})| \leq \xi
\end{equation}

\noindent the prediction $\sig{out}'$ that is computed using $\sig{in}'$ without perturbation (i.e., using inference with $\Xi = 0$), is guaranteed to have the following properties.
\vspace{-2mm}

\begin{enumerate}[label=(\alph*)]
    \item   If $\sig{lb}_{\langle i, j, 1\rangle} = 1$ then $\sig{out}'_{\langle 1, i, j, 1\rangle} \geq \alpha$. 
    \item If $\sig{lb}_{\langle i, j, 1\rangle} = 0$ then  $\sig{out}'_{\langle 1, i, j, 1\rangle} \leq \alpha$.  
    \item For $k \in \{2, \ldots, 7\}$,  $|\sig{out}'_{\langle 1, i, j, k\rangle} - \sig{lb}_{\langle i, j, k\rangle}| \leq \delta_k$.
\end{enumerate}

\end{lemma}   

\vspace{-5mm}

\paragraph{(Implementing the Loss Function)} As state-of-the-art neural network training frameworks such as TensorFlow or Pytorch always operate on batches of data, it is important that the previously mentioned loss functions can be implemented with batch support. Overall, the loss function defined in Definition~\ref{defi.loss} and~\ref{defi.robust.loss} is a  composition from two elements, namely (i) to apply Definition~\ref{defi.distance} and (ii) to perform case split depending on the value of the label, which  can only be~$1$ or~$0$. Table~\ref{table.implementation} details how to implement these two elements  in PyTorch.

\begin{table}[t]
\centering
\begin{tabular}{|c|c|}
\hline
Function on single input & Batched implementation in PyTorch  \\ \hline
$\sig{dist}(\sig{v}, \sig{low}, \sig{up})$  &   \texttt{torch.max(torch.clamp(low - v, min=0),} \\
& \texttt{torch.clamp(v - up, min=0))}  \\ \hline
If \sig{lb}=1 then x;  &   \texttt{torch.mul(\sig{lb}, x) + torch.mul(1 - \sig{lb}, y)} \\ 
else (i.e., \sig{lb}=0)  y  & \\
\hline
\end{tabular}
\vspace{1mm}
\caption{Implementing the loss function}
\label{table.implementation}
\vspace{-8mm}
\end{table}

\vspace{-3mm}
\section{Concluding Remarks}~\label{sec.concluding.remarks}
\vspace{-4mm}

In this paper, we exemplified how to extend a state-of-the-art single-stage 3D  detector neural network (\sig{PIXOR}) in a safety-aware fashion. By safety-aware, our goal is to reflect the safety specification into the architectural design, the engineering of the loss function, and the post-processing algorithm. Our proposed hardening is compatible with standard training methods while being complementary to other critical activities in the safety engineering of machine learning systems such as rigorous data collection, testing (for understanding the generalizability) or interpretation (for understanding the decision of networks). In our example, the tolerance concept integrated inside the loss function avoids unrealistic-and-unreachable perfection in training, allows integrating the idea of provable robustness, and finally, enables connecting the specification to capabilities or limitations from other components such as motion planners. 

For future work, we are interested in migrating the concept and the research prototype into real systems, as well as reflecting other safety specifications into the design of the architecture and the loss function. Yet another direction is to consider how other architectures used in multi-view 3D reconstruction (e.g., MV3D network~\cite{chen2017multi}) can also be made safety-aware.

%For future work, we plan to experiment the concept into the production system. 

\bibliographystyle{abbrv}

\newpage
\section*{Appendix (Proofs and Evaluation)}

%Contents in the appendix are for review purposes and will not be included in the final version. 

\subsection*{A. Proofs}

\setcounter{lemma}{0}

\begin{lemma}

For the labelled data $(\sig{in}, \sig{lb}) \in \mathcal{T}$, given $\sig{out} := f^{(n)}(\sig{in})$, let $\theta$ be the angle produced by $\sig{out}_{\langle i, j, 2\rangle}$ and $\sig{out}_{\langle i, j, 3\rangle}$. If $\sig{loss}_{\langle i, j\rangle}(\sig{out}, \sig{lb}) = 0$ and if $\sig{lb}_{_{\langle i, j, 1\rangle}} = 1$, then enlarging the prediction bounding box by 
\begin{itemize}
    \item enlarging the predicted width with $d + d_w$
    \item enlarging the predicted length with $d + d_l$
\end{itemize}
guarantees to contain the vehicle bounding box from label $\sig{lb}$ at grid ${\langle i, j\rangle}$, where 
\begin{itemize}
    \item $d > \sqrt{(\delta_4)^2 + (\delta_5)^2}$,

    \item  $d_l > \sig{max}_{\alpha \in [-\kappa, \kappa]}\;\frac{1}{2}(10^{\sig{out}_{\langle i, j, 6\rangle} + \delta_6} \sin \alpha + 10^{\sig{out}_{\langle i, j, 7\rangle} + \delta_7} \cos \alpha - 10^{\sig{out}_{\langle i, j, 7\rangle}}) $, 

\item  $d_w > \sig{max}_{\alpha \in [-\kappa, \kappa]}\;\frac{1}{2}(10^{\sig{out}_{\langle i, j, 6\rangle} + \delta_6} \cos \alpha + 10^{\sig{out}_{\langle i, j, 7\rangle} + \delta_7} \sin \alpha - 10^{\sig{out}_{\langle i, j, 6\rangle}})$,
\end{itemize}
and the interval $[-\kappa, \kappa]$ used in $d_l$ and $d_w$ satisfies the following constraints: 
\begin{multline}\label{eq.enlarge.buffer}
\forall \alpha \in [\frac{-\pi}{2},\frac{\pi}{2}]: (|\cos (\theta ) - \cos (\theta+\alpha)| \leq \delta_2 \; \wedge   |\sin (\theta ) - \sin (\theta+\alpha)| \leq \delta_3) \\ \rightarrow  \alpha \in [-\kappa, \kappa] 
\end{multline}
\end{lemma}

\proof As $\sig{lb}_{_{\langle i, j, 1\rangle}} = 1$, the label indicates the existence of vehicle at output grid $\langle i, j\rangle$. As $\sig{loss}_{\langle i, j\rangle}(\sig{out}, \sig{lb}) = 0$,  following the definition of $\sig{loss}_{\langle i, j\rangle}$ we obtain $\eta = 0$, meaning that for every $k \in [2, 7]$, $\sig{dist}(\sig{out}_{\langle i, j, k\rangle}, \sig{lb}_{\langle i, j, k\rangle} - \delta_k,  \sig{lb}_{\langle i, j, k\rangle} + \delta_k) = 0$, i.e., every $\sig{out}_{\langle i, j, k\rangle}$ only deviates by $\sig{lb}_{\langle i, j, k\rangle}$ with maximum amount $\delta_k$.

\vspace{1mm}

\begin{figure}[t]
%trim={<left> <lower> <right> <upper>}
    \centering
    \includegraphics[width=\textwidth, trim=2cm 5.2cm 2.5cm 7cm ]{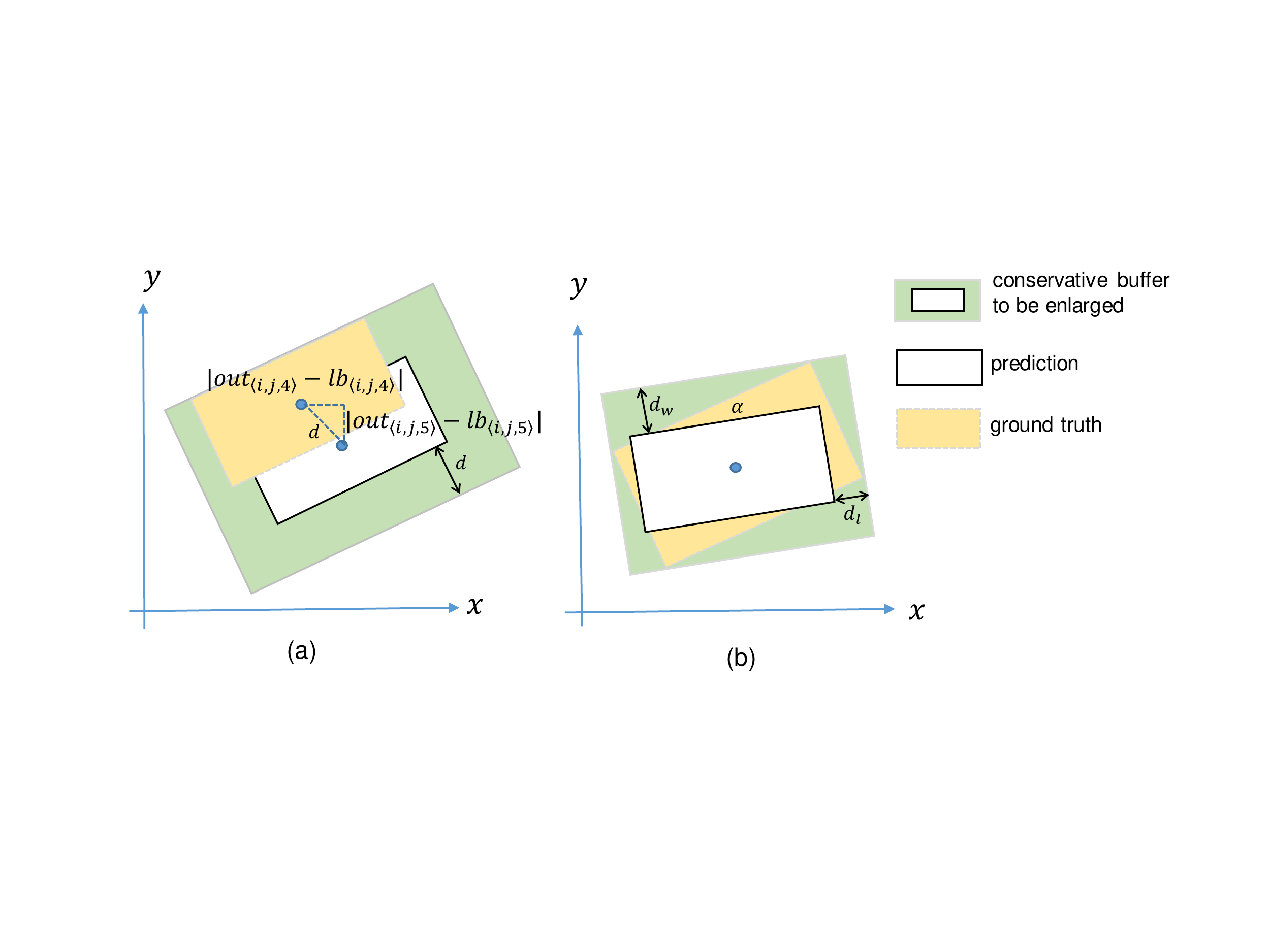}
    \caption{Understanding the required buffer size.}
    \label{fig:buffer.computation}
\end{figure}

To compute the required buffer, we consider (1) the effect of prediction deviation in terms of the center position of the vehicle  (i.e., $d_x$ for $\sig{out}_{_{\langle i, j, 4\rangle}}$ and $d_y$  for $\sig{out}_{_{\langle i, j, 5\rangle}}$), and (2) the effect of scaling and rotation. Effects (1) and (2) are independent, so the total required buffer is the sum of individual contributions from~(1) and~(2). An illustration can be found in Figure~\ref{fig:buffer.computation}.

\begin{itemize}
    \item For (1), as demonstrated in Figure~\ref{fig:buffer.computation}-a, the deviation in $d_x$, i.e., $|\sig{out}_{\langle i, j, 4\rangle} - \sig{lb}_{\langle i, j, 4\rangle}|$, is bounded by $\delta_4$ due to zero loss. Similarly, the deviation in $d_y$ is bounded by $\delta_5$. For the prediction to contain the bounding box, it suffices to enlarge both the length and the width by at least the distance of displacement, i.e., $d\geq \sqrt{(\delta_4)^2 + (\delta_5)^2}$. 
    
    \item For (2), under the condition of zero loss, it suffices to consider the length and the width of the ground truth to be the largest possible, i.e., in the ground truth, the length of the vehicle equals $L=10^{\sig{out}_{\langle i, j, 7\rangle} + \delta_7}$ and the width of the vehicle $W=10^{\sig{out}_{\langle i, j, 6\rangle} + \delta_6}$. Illustrated in Figure~\ref{fig:buffer.computation}-b, if the angle between the prediction and the ground truth equals~$\alpha$, 
    \begin{itemize}
   \item the length for the outer green rectangle equals $L\cos\alpha + W\sin \alpha$, and
   \item the width for the outer green rectangle equals $L\sin\alpha + W\cos \alpha$.
   \end{itemize}
   
   To find the size of the required buffer, we need to at least consider every possible angle deviation~$\alpha$ that still creates zero loss.  Equation~\ref{eq.enlarge.buffer} stated in the lemma provides such a characterization\footnote{The interval $[-\kappa, \kappa]$ is essentially a \emph{conservative upper bound on the deviated angle} between prediction and the ground truth, where their associated sine and cosine value differences are bounded by $\delta_2$ and $\delta_3$. As an example, if between  the predicted angle and the ground truth, we only allow the sine and cosine value to only differ by at most $0.1$ (i.e., $\delta_2=\delta_3=0.1$), it is easy to derive that $\kappa$ can be conservatively set to~$\frac{\pi}{18}$, i.e., ($10^{\circ}$), rather than the trivial value~$\frac{\pi}{2}$. This is because an angle difference of $10^{\circ}$ can already make sine and cosine value differ by~$0.15$, thereby creating non-zero loss. Therefore, conservatively setting $[-\kappa, \kappa]$ to be $[-\frac{\pi}{18}, \frac{\pi}{18}]$ in computing $d_l$ and $d_w$ surely covers all possible angle deviation constrained by zero loss.}.  
   
   Therefore, $d_l$, the length to be enlarged in the result bounding box under rotation and size expansion, should be larger than half of $L\cos\alpha + W\sin \alpha $ subtracted by the originally predicted size $10^{\sig{out}_{\langle i, j, 7\rangle}}$, which equals  \[\sig{max}_{\alpha \in [-\kappa, \kappa]}\;\frac{1}{2}(10^{\sig{out}_{\langle i, j, 6\rangle} + \delta_6} \sin \alpha + 10^{\sig{out}_{\langle i, j, 7\rangle} + \delta_7} \cos \alpha - 10^{\sig{out}_{\langle i, j, 7\rangle}}).\] 
   The computation for $d_w$ is analogous and is omitted here.
   
\end{itemize}
\qed

\begin{lemma}
Given $(\sig{in}, \sig{lb}) \in \mathcal{T}$, under the condition $\sig{robust\_loss}_{\langle i, j\rangle}(\sig{out}, \sig{lb}) = 0$ where $\sig{out}$ is computed using~$\sig{in}$ with perturbation bound $\Xi := \xi $ ($\xi> 0$) at layer~$n$, then for any $\sig{in}'$ where the following condition holds,
\begin{equation}~\label{eq.distance}
   \forall m \in \{1, \ldots, d_{(n-1)}\}: |g^{(n-1)}_{m}(\sig{in'}) - g^{(n-1)}_{m}(\sig{in})| \leq \xi
\end{equation}

\noindent the prediction $\sig{out}'$ that is computed using $\sig{in}'$ without perturbation (i.e., using inference with $\Xi = 0$), is guaranteed to have the following properties.

\begin{enumerate}[label=(\alph*)]
    \item   If $\sig{lb}_{\langle i, j, 1\rangle} = 1$ then $\sig{out}'_{\langle 1, i, j, 1\rangle} \geq \alpha$. 
    \item If $\sig{lb}_{\langle i, j, 1\rangle} = 0$ then  $\sig{out}'_{\langle 1, i, j, 1\rangle} \leq \alpha$.  
    \item For $k \in \{2, \ldots, 7\}$,  $|\sig{out}'_{\langle 1, i, j, k\rangle} - \sig{lb}_{\langle i, j, k\rangle}| \leq \delta_k$.
\end{enumerate}

\end{lemma}    

\proof (Sketch) Here we provide the proof for (a); for (b) and (c) the argument is analogous.
First we recall Assumption~\ref{assumption.exact} which states that when $\Xi$ is set to~$0$, the bound computation is exact and the computed lower- and upper-bound collide, i.e., for any input~$\sig{in}'$, we have $\sig{out}'_{\langle 1, i, j\rangle} = \sig{out}'_{\langle 2, i, j\rangle}$ and we can use~$\sig{out}'_{\langle 1, i, j\rangle}$ to be the result of prediction.

When an input~$\sig{in}'$ satisfies the constraint stated in Eq.~\ref{eq.distance}, the corresponding output $\sig{out}'_{\langle 1, i, j\rangle}$ where~$\Xi$ is set to~$0$, is contained in the lower- and upper-bound of $\sig{out}$ which is computed using $\sig{in}$ with~$\Xi$ being set to~$\xi$. As $\sig{robust\_loss}_{\langle i, j\rangle}(\sig{out}, \sig{lb})$ equals~$0$,  when $\sig{lb}_{\langle i, j, 1\rangle} = 1$ the predicted probability lower-bound is always larger or equal to~$\alpha$. Therefore, $\sig{out}'_{\langle 1, i, j, 1\rangle}$, being contained in the lower- and upper-bound, is also larger or equal to~$\alpha$.
\qed

\subsection*{B. Evaluation}

%Compare the area size between non-max-suppression and non-max-inclusion. 
In this section, we provide a summary of our initial evaluation under \textsf{PIXOR} and the KITTI dataset, following the original paper~\cite{yang2018pixor}. 
The evaluation using KITTI dataset in this paper is for knowledge dissemination and scientific publication and is not for commercial use.

\begin{figure}[t]
%trim={<left> <lower> <right> <upper>}
\centering
\includegraphics[width=\textwidth, trim=2.5cm 1cm 3.5cm 3cm, clip]{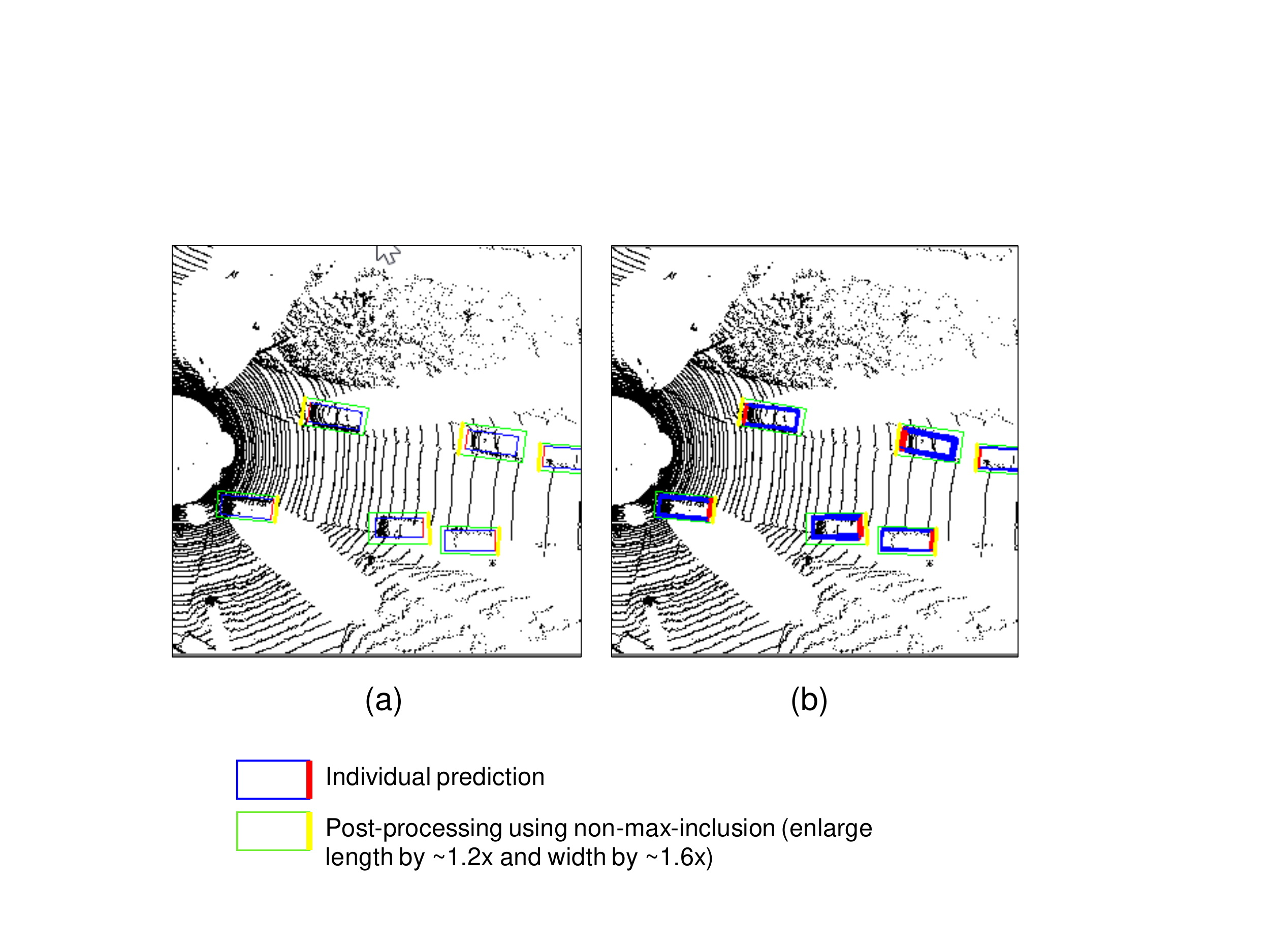}
\caption{Using non-max-inclusion with \sig{PIXOR}.}
\label{fig:kitti.non.max.inclusion}
\end{figure}

\paragraph{(Understanding the effect of non-max-inclusion)} We have implemented the non-max-suppression algorithm to understand the effect in KITTI dataset. To create a bounding box that contains all predictions, in our prototype implementation (admittedly non-optimal), we enlarge the length of the bounding box from $l_{box}$ to $l_{box} (1.05)^i$ and the width from $w_{box}$ to $w_{box} (1.1)^i$, where~$i$ is iterated from~$1$ until all related prediction 
can be included. In Figure~\ref{fig:kitti.non.max.inclusion}-a, one sees the  bounding boxes with the largest prediction probability (blue) and their enlarged version (in green). Figure~\ref{fig:kitti.non.max.inclusion}-b demonstrates all predicted bounded boxes that are being included; as multiple blue rectangles are imperfectly overlaid due to variations in size and position, the resulting blue lines looks thicker. As for \sig{PIXOR} the grid size is actually very small ($0.4 \times 0.4 m^2$), for predicting a single vehicle there can be up to~$20$ bounding boxes that require to be included. For the enlarged bounding box of a vehicle, the width may require a~$60\%$ increase, while the length may require a~$30\%$ increase.

\paragraph{(New loss function with tolerance and robust training)} We have conducted an initial experiment for understanding the effect of the new loss function and robust training. In the following, we explain in detail how the evaluation is conducted. 

We use the open-source version of \sig{PIXOR} available at the following website \texttt{https://github.com/ankita-kalra/PIXOR} 
as the baseline for experimenting the concept. In the original implementation, the geometrical information regarding the input and output is stored in the data structure \texttt{
geometry = \{'L1': -40.0, 'L2': 40.0, 'W1': 0.0, 'W2': 70.0, 'H1': -2.5, 'H2': 1.0,\\ 'input\_shape': (800, 700, 36),'label\_shape': (200, 175, 7)\}}. We have created the following modified configuration to consider the critical area. 
 \texttt{geometry = \{'L1': -15.0, 'L2': 15.0, 'W1': 0.0, 'W2': 30.0, 'H1': -2.5, 'H2': 1.0, 'input\_shape': (300, 300, 36), 'label\_shape': (75, 75, 7)\}}. The considered critical area only produces a 2D output grid with size $75\times75$, apart from the original setup that produces $200 \times 175$. 
This also influences the decoder from the generated result, where \texttt{self.geometry} (for interpreting the output) is also correspondingly set to \texttt{[-15.0, 15.0, 0.0, 30.0]} (for $x$ and $y$ coordinate, it spans from $-15$~m to $15$~m and from $0$~m to $30$~m. With a grid size of $0.4$, it creates $\frac{15-(-15)}{0.4} \times \frac{30-(0)}{0.4}= 75\times 75$ in the output grid, matching the \texttt{'label\_shape'} attribute). This overall enables a training using a single GPU with $11$~GB memory, operated with a batch of~$24$.  

For the robust training, in the network the symbolic perturbation is enabled at~$3$ layers before the output is produced. We allow using batch normalization before the layer where symbolic error is introduced. The following code snippet demonstrates the required modification for the header network for forward reasoning, where a perturbation bound $\Xi=0.001$ is used in training. One can integrate this code snippet to the network and use correspondingly the post-processing algorithm. 

\begin{small}
\begin{verbatim}

def forward(self, x):

    xi = 0
    if self.training :
       xi = 0.001  

    x = self.conv1(x)
    if self.use_bn:
        x = self.bn1(x)
    x = self.conv2(x)
    if self.use_bn:
        x = self.bn2(x)
    
    # Add the perturbation bound such that value changes to 
    # [l_o2, u_o2] = [x - xi,  x + xi]. 
    
    # When xi = 0, l_o2 = u_o2, and the computation below ensures that 
    # l_o3 = u_o3, and l_o4 = u_o4
    
    l_o2 = x - xi    
    u_o2 = x + xi  
    
    # ReLU over conv3, symbolic version
    l_o3 = (nn.functional.conv2d(l_o2, self.conv3.weight.clamp(min=0), bias=None, 
                stride=self.conv3.stride, padding=self.conv3.padding,
                dilation=self.conv3.dilation, groups=self.conv3.groups) +
           nn.functional.conv2d(u_o2, self.conv3.weight.clamp(max=0), bias=None, 
                stride=self.conv3.stride, padding=self.conv3.padding,
                dilation=self.conv3.dilation, groups=self.conv3.groups) +
            self.conv3.bias[None,:,None,None])
    
    u_o3 = (nn.functional.conv2d(u_o2, self.conv3.weight.clamp(min=0), bias=None, 
                    stride=self.conv3.stride, padding=self.conv3.padding,
                    dilation=self.conv3.dilation, groups=self.conv3.groups) +
              nn.functional.conv2d(l_o2, self.conv3.weight.clamp(max=0), bias=None, 
                    stride=self.conv3.stride, padding=self.conv3.padding,
                    dilation=self.conv3.dilation, groups=self.conv3.groups) + 
              self.conv3.bias[None,:,None,None])  
    
    # ReLU over conv4, symbolic version 
    l_o4 = (nn.functional.conv2d(l_o3, self.conv4.weight.clamp(min=0), bias=None, 
                    stride=self.conv4.stride, padding=self.conv4.padding,
                    dilation=self.conv4.dilation, groups=self.conv4.groups) +
            nn.functional.conv2d(u_o3, self.conv4.weight.clamp(max=0), bias=None, 
                    stride=self.conv4.stride, padding=self.conv4.padding,
                    dilation=self.conv4.dilation, groups=self.conv4.groups) +
            self.conv4.bias[None,:,None,None])

    u_o4 = (nn.functional.conv2d(u_o3, self.conv4.weight.clamp(min=0), bias=None, 
                    stride=self.conv4.stride, padding=self.conv4.padding,
                    dilation=self.conv4.dilation, groups=self.conv4.groups) +
            nn.functional.conv2d(l_o3, self.conv4.weight.clamp(max=0), bias=None, 
                    stride=self.conv4.stride, padding=self.conv4.padding,
                    dilation=self.conv4.dilation, groups=self.conv4.groups) + 
            self.conv4.bias[None,:,None,None])           

    if self.training == True :
    # In training mode, produce min and max 
    # (will be further processed by post-processor)
        cls_min = torch.sigmoid(self.clshead(l_o4))
        reg_min = self.reghead(l_o4)
        cls_max = torch.sigmoid(self.clshead(u_o4))
        reg_max = self.reghead(u_o4)
        return cls_min, reg_min, cls_max, reg_max
    
    else:
    # In inference mode, as xi is set to 0, just use lower bound as result
        cls = torch.sigmoid(self.clshead(l_o4))
        reg = self.reghead(l_o4)
        return cls, reg
        
\end{verbatim}

\end{small}

We started with a network trained using the loss defined in~\cite{yang2018pixor} that is also implemented in the above mentioned public github repository. One observes that for some models, when setting  $\Xi = 0.001$, $\alpha = 0.5$ (class threshold) and $\delta_k = 0.01$ ($k \in \{2, \ldots, 7\}$, tolerance in each prediction), the computed $\sig{robust\_loss}$ is already~$0$. This means that the network has an inherent tolerance of over perturbation, if the amount of adversarial perturbation reflected on the feature-level is very small ($\Xi = 0.001$). Here we omit further explanations, but one can surely increase the value of~$\Xi$ in order to adjust the parameters of the network to be more robust against perturbation. 

\end{document}